\documentclass[10pt,twocolumn,letterpaper]{article}

 \usepackage{cvpr}              % To produce the CAMERA-READY version
\usepackage{graphicx}
\usepackage{amsmath}
\usepackage{amssymb}
\usepackage{booktabs}
\usepackage[accsupp]{axessibility}
\usepackage{xcolor}
\newcommand*{\affaddr}[1]{#1} % No op here. Customize it for different styles.
\newcommand*{\affmark}[1][*]{\textsuperscript{#1}}

\usepackage[pagebackref,breaklinks,colorlinks]{hyperref}

\usepackage{bbold}
\usepackage{multirow}
\usepackage{bm}
\usepackage{arydshln}
\usepackage{xspace}
\usepackage{url}

\makeatletter
\newcommand{\figcaption}[1]{\def\@captype{figure}\caption{#1}}
\newcommand{\tblcaption}[1]{\def\@captype{table}\caption{#1}}
\newcommand\customparagraph[1]{\vspace{0.7em}\noindent\textbf{#1}}
\makeatother

\DeclareRobustCommand\green{\color[rgb]{0,0.5,0}}

\def\tbf{\textbf}

\makeatletter
\DeclareRobustCommand\onedot{\futurelet\@let@token\@onedot}
\def\@onedot{\ifx\@let@token.\else.\null\fi\xspace}
\def\eg{\emph{e.g}\onedot} 
\def\ie{\emph{i.e}\onedot} 
 
 \def\vs{\emph{vs}\onedot}
 
\def\etal{\emph{et al}\onedot}

\makeatother
\usepackage[capitalize]{cleveref}
\crefname{section}{Sec.}{Secs.}
\Crefname{section}{Section}{Sections}
\Crefname{table}{Table}{Tables}
\crefname{table}{Tab.}{Tabs.}

\definecolor{C0}{RGB}{255,0,0}
\definecolor{C1}{RGB}{0,160,250}
\definecolor{C2}{RGB}{200,0,0}
\definecolor{C3}{RGB}{150,150,0}
\definecolor{C4}{RGB}{50,50,255}

\newcommand{\volnet}{MVExoNet}
\newcommand{\svnet}{SVEgoNet}
\newcommand{\nego}{{490K}}
\newcommand{\ntotal}{{3.0M}}

\newcommand{\fpsar}{{30}}
\def\cvprPaperID{5180} % *** Enter the CVPR Paper ID here
\def\confName{CVPR}
\def\confYear{2023}

\def\tbldata{
\begin{table*}[tp]
\small{
\begin{tabular}{l|crrrrl}
Dataset                             & Modality                  & \multicolumn{1}{c}{\#img} & \multicolumn{1}{l}{\#ego\_img} & \multicolumn{1}{l}{\#views}  & \multicolumn{1}{l}{\#subj} & Annotation approach                                                                                    \\ \hline\hline
EgoDexter~\cite{mueller:iccv17}     & RGB-D                     & 3K                        & 3K                            & 1 (ego)                      & 4                          & Manual                                                                                                 \\
Panoptic Studio~\cite{simon:cvpr17} & RGB                       & 15K                       & -                             & 31                           & N/A                        & 2D + triangulation                                                                                     \\
FPHA~\cite{hernando:cvpr18}         & RGB-D                     & 105K                      & 105K                          & 1 (ego)                      & 6                          & Magnetic sensor                                                                                        \\
FreiHAND~\cite{zimmermann:iccv19}   & RGB                       & 37K                       & -                             & 8                            & 32                         & Manual + 3D volume + template fitting                                                                  \\
HO3D~\cite{hampali:cvpr20}          & RGB-D                     & 103K                      & -                             & 5                            & 10                         & 2D + template fitting                                                                                  \\
InterHand2.6M~\cite{moon:eccv20}    & RGB                       & 2.59M                      & -                             & 80-140                       & 27                         & Manual + 2D + triangulation                                                                            \\
DexYCB~\cite{chao:cvpr21}           & RGB-D                     & 508K                      & -                             & 8                            & 10                         & Manual + template fitting                                                                              \\
H2O~\cite{kwon:iccv21}              & RGB-D                     & 571K                      & 114K                          & 4 + 1 (ego)                  & 4                          & 2D + template fitting + smoothing                                                                      \\ \hline
AssemblyHands (M)                   & \multirow{3}{*}{RGB/Mono} & 227K                      & 22K                           & \multirow{3}{*}{8 + 4 (ego)} & 14                         & \multirow{3}{*}{\begin{tabular}[c]{@{}l@{}}
Manual + 3D volume + refinement
\end{tabular}} \\
AssemblyHands (A)                   &                           & 2.81M                       & 468K                         &                              & 20                         &                                                                                                        \\
\tbf{AssemblyHands (M + A)}         &                           & 3.03M                       & 490K                         &                              & 34                         &                            % AssemblyHands (Ours)                & Mono/RGB & 400K      & 66.8K         &  8 + 4 ego  & 10        & Manual + volumetric + refinement + smoothing
\end{tabular}
\caption[Caption for LOF]{
\textbf{Comparison of AssemblyHands with existing 3D hand pose datasets.\footnotemark[1]}
``M'' and ``A'' stand for manual and automatic annotation, respectively.
AssemblyHands is the largest existing benchmark for egocentric 3D hand pose estimation.
}
\label{table:dataset}
}
\end{table*}
}

\def\tblannot{
\begin{table}[t]
\centering
\begin{tabular}{l|cc}
Annotation method                                                                   & MPJPE & PCK-AUC \\ \hline \hline
Egocentric-only~\cite{sener:cvpr22}                                  & 27.55 & 29.4    \\ \hdashline
2D + Triangulation                                                                  & 7.97  & 63.8    \\ %\hline
\volnet-R1 (Ours)                                                             & 5.42  & 79.2     \\
\volnet-R2 (Ours)                                                             & 4.30  & 83.1      \\
\volnet-R3 (Ours)                                                             & \tbf{4.20}  & \tbf{83.4}      % \\
\end{tabular}
\caption{\textbf{Evaluation of hand pose annotation on manually annotated subset of AssemblyHands.}
We use MPJPE (mm) and  PCK-AUC (\%) as the evaluation metrics.
}
\label{table:annot}
\end{table}
}

\def\tblegohoi{
\begin{table}[t]
\centering
\begin{tabular}{l|cc}
Annotation method                                                                   & MPJPE & PCK-AUC \\ \hline \hline
2D + Triangulation                                                                  & 49.21 & 23.9    \\ %\hline
\volnet-R1 (Ours)                                                             & 21.20 & 51.3    \\
\volnet-R2 (Ours)                                                             & 14.57 & 67.2    \\
\volnet-R3 (Ours)                                                             & \tbf{13.38} & \tbf{70.4}    
\end{tabular}
\caption{\textbf{Evaluation of multi-view annotation on the Desktop Activities dataset~\cite{lv_aria:2022}.} 
We use MPJPE (mm) and  PCK-AUC (\%) as the evaluation metrics.
}
\label{tbl:egohoi}
\end{table}
}

\def\tblcross{
\begin{table}[t]
\centering
\begin{tabular}{l|c:cc}
Subsets   & \multicolumn{1}{l:}{Eval-M} & \multicolumn{1}{l}{Eval-A} & \multicolumn{1}{l}{Eval-M+A} \\ \hline  \hline
Train-M   & 24.38                        & 28.58                        & 28.35                            \\
Train-A   & 25.18                        & 22.29                        & 22.45                                \\
Train-M+A & \tbf{23.46}                  & \tbf{21.84}                  & \tbf{21.92}                
\end{tabular}
\caption{
\textbf{
Effect of automatic annotation for the training of \svnet{}.}
We use egocentric image sets with manual (M), automatic (A), and manual and automatic (M + A) annotation for training and evaluation.
We report MPJPE (mm) as the evaluation metric (lower is better).
}
\label{table:cross}
\end{table}
}

\def\tblego{
\begin{table*}[t]
\centering
\begin{tabular}{l|c|ccccccc}
Method   & \multicolumn{1}{l|}{MPJPE} %& \multicolumn{1}{l}{PCK-AUC} 
& pick up & put down & position & remove  & screw & unscrew 
& \multicolumn{1}{l}{Avg. Verb Acc.} 
\\ \hline\hline
UmeTrack~\cite{dpe} & 32.91  & 66.0 & 41.5 & 51.2 & \tbf{29.2} & 42.5 & 59.6 & 50.3  (83.8\%) \\ %\hdashline 
\svnet{} (Ours)        & \tbf{21.92}  & \tbf{68.1}  &  \tbf{47.7} &  \tbf{64.3} &  27.4 &  \tbf{44.3} &  \tbf{63.7} &  \tbf{54.7} (91.1\%)    \\ \hline
AssemblyHands-A  & - & 70.0 & 57.4 & 67.5 & 36.4 & 49.8 & 64.1 & 60.0 (100\%)
\end{tabular}
\caption{
\textbf{
{Evaluation of action classification from hand poses.}}
We train and evaluate a MS-G3D~\cite{liu2020disentangling} action classification model using hand pose sequences as input, and report Verb Accuracy (\%).
AssemblyHands-A represents the empirical upper bound where automatically annotated hand poses are used as input.
Our \svnet{} predicts more accurate 3D hand poses, which leads to better classification accuracy.
}
\label{table:ego}
\end{table*}
}
\begin{document}

\title{
AssemblyHands:\\ Towards Egocentric Activity Understanding via 3D Hand Pose Estimation
}

\newcommand{\namesep}{\hspace{0.8em}}
\newcommand{\linesep}{\vspace{0.4em}}
\author{Takehiko Ohkawa\affmark[1,2]\footnotemark\namesep
Kun He\affmark[1]\namesep
Fadime Sener\affmark[1]\namesep
Tomas Hodan\affmark[1]\namesep
Luan Tran\affmark[1]\namesep
Cem Keskin\affmark[1] \linesep \\
\affaddr{\affmark[1]Meta Reality Labs\namesep\affmark[2]The University of Tokyo} \linesep \\
% {\email{\{tohkawa,kunhe,famesener,tomhodan,tranluan07,cemkeskin\}@meta.com}}\\
% Project page:
\href{https://assemblyhands.github.io/}{https://assemblyhands.github.io}
}
\maketitle
\footnotetext[0]{*~Work done during internship.}

%%%%%%%%%%%%%%%%%%%%%%%%%%%%%%%%%%%%%%%%%%%%%%%%%%%%%%%%%%%%%%%
%%%%%%%%%%%%%%%%%%%%%%%%%%%%%%%%%%%%%%%%%%%%%%%%%%%%%%%%%%%%%%%
%%%%%%%%% ABSTRACT
%%%%%%%%%%%%%%%%%%%%%%%%%%%%%%%%%%%%%%%%%%%%%%%%%%%%%%%%%%%%%%%
%%%%%%%%%%%%%%%%%%%%%%%%%%%%%%%%%%%%%%%%%%%%%%%%%%%%%%%%%%%%%%%
\begin{abstract}
We present \tbf{AssemblyHands}, a large-scale benchmark dataset with accurate 3D hand pose annotations,
to facilitate the study of egocentric activities with challenging hand-object interactions.
The dataset includes synchronized egocentric and exocentric images sampled from the recent Assembly101 dataset, in which participants assemble and disassemble take-apart toys.
To obtain high-quality 3D hand pose annotations for the egocentric images, 
we develop an efficient pipeline, where we use an initial set of manual annotations to train a model to automatically annotate a much larger dataset.
Our annotation model uses multi-view feature fusion and an iterative refinement scheme,
and achieves an average keypoint error of 4.20\,mm, which is {85\%} lower than the error of the original annotations in Assembly101.
AssemblyHands provides \ntotal{} annotated images, including \nego{} egocentric images, making it the largest existing benchmark dataset for egocentric 3D hand pose estimation.
Using this data, we develop a strong single-view baseline of 3D hand pose estimation from egocentric images.
Furthermore, we design a novel action classification task to evaluate predicted 3D hand poses.
Our study shows that having higher-quality hand poses directly improves the ability to recognize actions.
\end{abstract}

%%%%%%%%%%%%%%%%%%%%%%%%%%%%%%%%%%%%%%%%%%%%%%%%%%%%%%%%%%%%%%%
%%%%%%%%%%%%%%%%%%%%%%%%%%%%%%%%%%%%%%%%%%%%%%%%%%%%%%%%%%%%%%%
%%%%%%%%% BODY TEXT
%%%%%%%%%%%%%%%%%%%%%%%%%%%%%%%%%%%%%%%%%%%%%%%%%%%%%%%%%%%%%%%
%%%%%%%%%%%%%%%%%%%%%%%%%%%%%%%%%%%%%%%%%%%%%%%%%%%%%%%%%%%%%%%
\section{Introduction}\label{sec:intro}

\begin{figure}[t]
    \centering
     \includegraphics[width=1\linewidth]{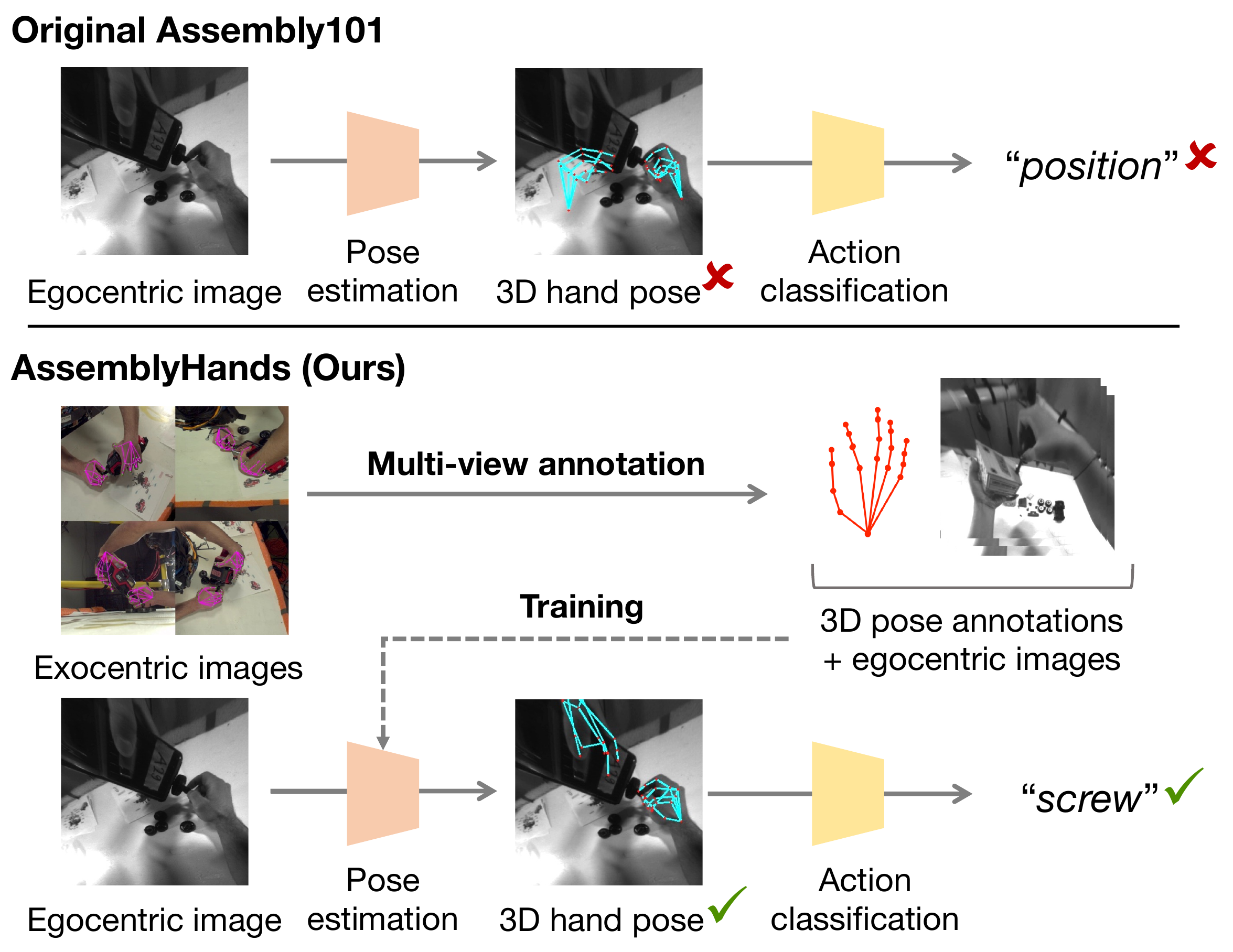}
    \caption{
    \tbf{High-quality 3D hand poses as an effective representation for egocentric activity understanding.}
    AssemblyHands provides high-quality 3D hand pose annotations computed from multi-view exocentric images sampled from Assembly101~\cite{sener:cvpr22}, which originally comes with inaccurate annotations computed from egocentric images (see the incorrect left-hand pose prediction).
    As we experimentally demonstrate on an action classification task, models trained on high-quality annotations achieve significantly higher accuracy.
    }
    \label{fig:teaser}
\end{figure}

% P1: Our motivation: egocentric HPE for AR/VR
Recognizing human activities  %(\eg cutting food, shooting a basketball) 
is a decades-old problem in computer vision~\cite{kong:2022ijcv}.
With recent advancements in user-assistive augmented reality and virtual reality (AR/VR) systems,
there is an increasing demand for recognizing actions from the \emph{egocentric} (first-person) viewpoint.
Popular AR/VR headsets such as Microsoft HoloLens, Magic Leap, and Meta Quest are typically equipped with egocentric cameras to capture a user's interactions with the real or virtual world.
In these scenarios, the user's hands manipulating objects is a very important modality of interaction.
In particular, hand poses (\eg, 3D joint locations) play a central role in understanding and enabling hand-object interaction~\cite{kwon:iccv21,chao:cvpr21}, pose-based action recognition~\cite{sener:cvpr22,hernando:cvpr18,liu2020disentangling}, and interactive interfaces~\cite{megatrack,dpe}.

\begin{figure*}[t]
    \centering
    \includegraphics[width=.9\linewidth]{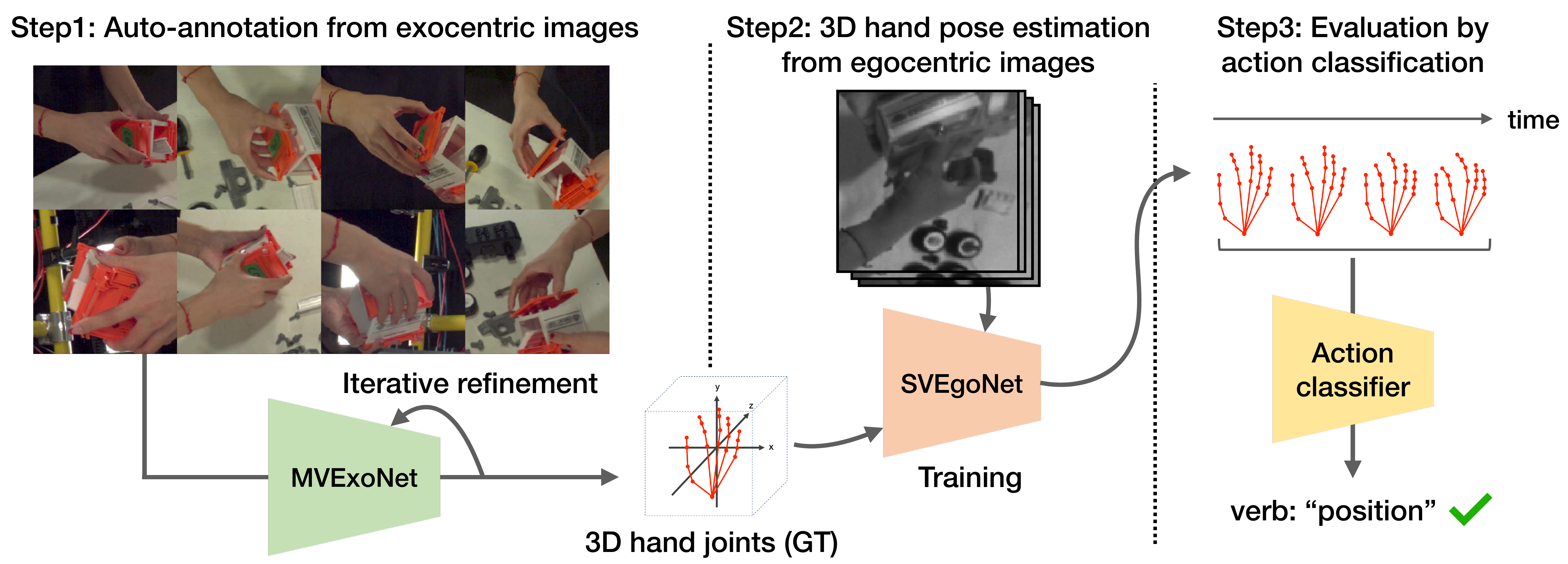}
    \caption{\tbf{Construction of AssemblyHands dataset and a benchmark task for egocentric 3D hand pose estimation.}
    We first use manual annotations and an automatic annotation network (MVExoNet) to generate accurate 3D hand poses for multi-view images sampled from the Assembly101 dataset~\cite{sener:cvpr22}.
    These annotations are used to train a single-view 3D hand pose estimation network (SVEgoNet) from egocentric images.
    Finally, the predicted hand poses are evaluated by the action classification task.
    }
    \label{fig:pipeline}
\end{figure*}

% P2: assembly intro
Recently, several large-scale datasets for understanding egocentric activities have been proposed, 
such as EPIC-KITCHENS~\cite{damen:ijcv21}, Ego4D~\cite{grauman:cvpr22}, and Assembly101~\cite{sener:cvpr22}.
In particular, Assembly101 highlights the importance of 3D hand poses in recognizing procedural activities such as assembling toys.
3D hand poses are compact representations, and are highly indicative of actions and even the objects that are interacted with-- 
for example, the ``screwing" hand motion is a strong cue for the presence of a screwdriver.
Notably, the authors of Assembly101 found that, for classifying assembly actions, learning from 3D hand poses is more effective than solely using video features.
However, a drawback of this study is that the 3D hand pose annotations in Assembly101 are not always accurate, as they are computed from an off-the-shelf egocentric hand tracker~\cite{dpe}.
We observed that the provided poses are often inaccurate (see Fig.~\ref{fig:teaser}), especially when hands are occluded by objects from the egocentric perspective.
Thus, the prior work has left us with an unresolved question: \emph{How does the quality of 3D hand poses affect action recognition performance?}

% P3: our work
To systematically answer this question, we propose a new benchmark dataset named \textbf{AssemblyHands}. 
It includes a total of \ntotal{} images sampled from Assembly101, annotated with high-quality 3D hand poses.
We not only acquire manual annotations, but also use them to train an accurate automatic annotation model that uses multi-view feature fusion from exocentric (\ie, third-person) images; please see Fig.~\ref{fig:pipeline} for an illustration.
Our model achieves 4.20\,mm average keypoint error compared to manual annotations, which is 85\% lower than the original annotations provided in Assembly101.
This automatic pipeline enables us to efficiently scale annotations to \nego{} egocentric images from 34 subjects, making AssemblyHands the largest egocentric hand pose dataset to date, both in terms of scale and subject diversity.
Compared to recent hand-object interaction datasets, such as DexYCB~\cite{chao:cvpr21} and H2O~\cite{kwon:iccv21}, our AssemblyHands features significantly more hand-object combinations, as each multi-part toy can be disassembled and assembled at will,

Given the annotated dataset, we first develop a strong baseline for egocentric 3D hand pose estimation, using 2.5D heatmap optimization and hand identity classification.
Then, to evaluate the effectiveness of predicted hand poses, we propose a novel evaluation scheme: action classification from hand poses.
Unlike prior benchmarks on egocentric hand pose estimation~\cite{mueller:iccv17,hernando:cvpr18,kwon:iccv21}, we offer detailed analysis of the quality of 3D hand pose annotation, its influence on the performance of an egocentric pose estimator, and the utility of predicted poses for action classification.

Our contributions are summarized as follows:
\begin{itemize}
    \vspace{-5pt}
    \setlength{\parskip}{0pt}
    \setlength{\itemsep}{3pt}
    \item We offer a large-scale benchmark dataset, dubbed AssemblyHands, with 3D hand pose annotations for \ntotal{} images sampled from the Assembly101 dataset, including \nego{} egocentric images.
    \item We propose an automatic annotation pipeline with multi-view feature fusion and iterative refinement, leading to 85\% error reduction in the hand pose annotations.
    \item We define a benchmark task for egocentric 3D hand pose estimation with the evaluation from action classification. 
    We provide a strong single-view baseline that optimizes 2.5D keypoint heatmaps and classifies hand identity.
    Our results confirm that having high-quality 3D hand poses significantly improves egocentric action recognition performance.
\end{itemize}

\tbldata

%%%%%%%%%%%%%%%%%%%%%%%%%%%%%%%%%%%%%%%%%%%%%%%%%%%%%%%%%%%%%%%
%%%%%%%%%%%%%%%%%%%%%%%%%%%%%%%%%%%%%%%%%%%%%%%%%%%%%%%%%%%%%%%
\section{Related work}\label{sec:related}

\customparagraph{Recognizing actions from pose.}
The general framework for recognizing people's actions involves extracting low-level states from sensor observations, such as image features or body/hand motion, and then feeding a temporal sequence of states into a recognition model.
There is a long history of using full body pose as the state representation in recognizing actions \cite{iqbal:2017,cheron:2015iccv,yao:2011bmvc,wang:2013cvpr,shahroudy2016ntu}, 
since poses are compact representations that contain discriminative information about actions. 
Also, in the context of AR/VR, pose information carries the benefit that its availability is less affected by privacy concerns, unlike image/video data. 
On the modeling side, graph convolutional networks, which treat joints as nodes and bones as edges, have been commonly used in skeleton-based action recognition~\cite{yu2017spatio,liu2020disentangling}.

In the exocentric setting, action recognition from hand poses is less explored compared to using full body pose, and is only studied on rather small datasets~\cite{kwon:iccv21}. 
Instead, hand poses are much  more relevant in the egocentric setting.
Recently, a large-scale dataset, Assembly101~\cite{sener:cvpr22}, was proposed to investigate action recognition using 3D hand poses.    
For Assembly101, 3D hand poses were found to be strong predictors of action;  
in particular, using hand poses was shown to give higher action classification accuracy compared to using video-based features~\cite{lin2019tsm}.

\footnotetext[1]{We do not include Assembly101 in Table~\ref{table:dataset} because it was not intended as a 3D hand pose dataset, and the provided poses do not have sufficient quality for training pose estimation models.}
\customparagraph{Datasets for 3D hand pose estimation.}
Table~\ref{table:dataset} shows statistics on existing RGB-based 3D hand pose datasets and our AssemblyHands.
Prior works on egocentric hand pose estimation annotate 2D keypoints on a depth image~\cite{mueller:iccv17} or use magnetic markers attached to hands~\cite{hernando:cvpr18}.
Due to the noise from these sensors, as well as the annotation cost, the accuracy and amount of annotation in these benchmarks are not sufficient.
Thus, most 3D hand pose estimation works focus on using inputs from static exocentric cameras~\cite{sridhar:eccv16,zimmermann:iccv17,hasson:cvpr19,simon:cvpr17,zimmermann:iccv19,hampali:cvpr20,moon:eccv20,chao:cvpr21} or utilize such an exocentric dataset to improve egocentric hand pose prediction~\cite{ohkawa:eccv22}.

Setups with multiple static cameras have several advantages and have been widely used in the literature~\cite{ohkawa:arxiv22}.
First, the total number of available images proportionately increases with the number of cameras.
For instance, InterHand2.6M~\cite{moon:eccv20} features numerous camera views (80+), resulting in the largest existing hand pose estimation dataset (non-egocentric) with a moderate amount of distinct frames.
Second, 3D keypoint coordinates can be reliably annotated from multiple 2D keypoints by using triangulation~\cite{simon:cvpr17,moon:eccv20} or hand template fitting~\cite{zimmermann:iccv19,hampali:cvpr20,chao:cvpr21,kwon:iccv21} (\eg, MANO~\cite{romero:tog17}).

Recently, a few egocentric activity datasets have installed synchronized egocentric cameras along with exocentric cameras, \eg, Assembly101~\cite{sener:cvpr22} and H2O~\cite{kwon:iccv21}.
The availability of exocentric images can significantly reduce the amount of annotation effort required for egocentric images.
Compared to the H2O dataset, AssemblyHands provides more than four times egocentric images with accurate ground truth and eight times the number of subjects.
With our higher sampling rate at \fpsar\,Hz, the total number of both egocentric and exocentric images (3.0M) surpasses the size of the InterHands2.6M.
Due to the goal-oriented nature of assembly actions, the hand poses in our benchmark are totally natural and unscripted, which is less focused in the existing study.
 
For automatic annotation, we utilize a volumetric convolution network similar to the one used by Zimmermann~\etal~\cite{zimmermann:iccv19}.
We further augment our model with an iterative refinement scheme to improve its accuracy, that does not require additional training.

%%%%%%%%%%%%%%%%%%%%%%%%%%%%%%%%%%%%%%%%%%%%%%%%%%%%%%%%%%%%%%%
%%%%%%%%%%%%%%%%%%%%%%%%%%%%%%%%%%%%%%%%%%%%%%%%%%%%%%%%%%%%%%%
\section{AssemblyHands dataset generation}\label{sec:dataset}

The input data in our proposed benchmark comes from the recently introduced Assembly101~\cite{sener:cvpr22}, a large-scale multi-view video dataset designed for understanding procedural activities, in particular, the assembly and disassembly of take-apart toys.
It is recorded with a static rig of 8 RGB cameras, plus 4 monochrome cameras on a synchronized headset worn by the human subject.

The initial hand pose annotations for Assembly101 are generated using an off-the-shelf hand tracker specifically designed for monochrome egocentric images~\cite{dpe}.
While it can estimate 3D hand poses with reasonable accuracy, there are several limitations.
For example, since the stereo area of the egocentric cameras is relatively narrow, depth estimates become inaccurate as hands move further away from the image center. 
Also, egocentric-only tracking is prone to severe failure modes due to heavy occlusion during hand-object interaction.
These motivate us to develop a multi-view annotation method using exocentric RGB cameras. 

While several existing datasets use off-the-shelf RGB-based models (\eg, OpenPose~\cite{cao:tpami2019}) to annotate hand poses, we have observed their accuracy is not satisfactory in Assembly101 (see the supplement for details). 
Since the OpenPose is trained on images with less hand-object occlusions~\cite{simon:cvpr17}, its predictions are often noisy when novel real-world objects (take-apart toys) and higher levels of occlusion 
are presented in Assembly101.
Thus, it is necessary to develop an annotation method tailored to our novel setup.

\subsection{Automatic annotation pipeline}
We present our proposed automatic annotation pipeline using multi-view exocentric RGB images.
We first prepare manual annotation for the frames sampled from the subset of Assembly101 at 1\,Hz.
Since obtaining manual annotations is laborious, we use them for training an annotation network that can automatically provide reasonable 3D hand pose annotation.
We then introduce the detail of our annotation network: (1) an annotation network using volumetric feature fusion (\volnet), and (2) iterative refinement during inference of the network.
Compared to the manual annotation, this automatic annotation scheme allows us to assign 21 times more labels in another subset of Assembly101 sampled at \fpsar\,Hz.

\customparagraph{Manual annotation.}
First, we obtain manual annotations of the 3D locations of 21 joints on both hands in the world coordinate space.
We use a setup similar to that of \cite{moon:eccv20,feng:arxiv21}, where 2D keypoints are annotated from multiple views and triangulated into 3D.
In total, we annotated 62 video sequences from Assembly101 at a sampling rate of 1 Hz, resulting in an annotated set of 22K frames, each having 8 RGB views.
We further split it into 54 sequences for training and 8 sequences for testing.

\begin{figure}[t]
    \centering
    \includegraphics[width=1\linewidth]{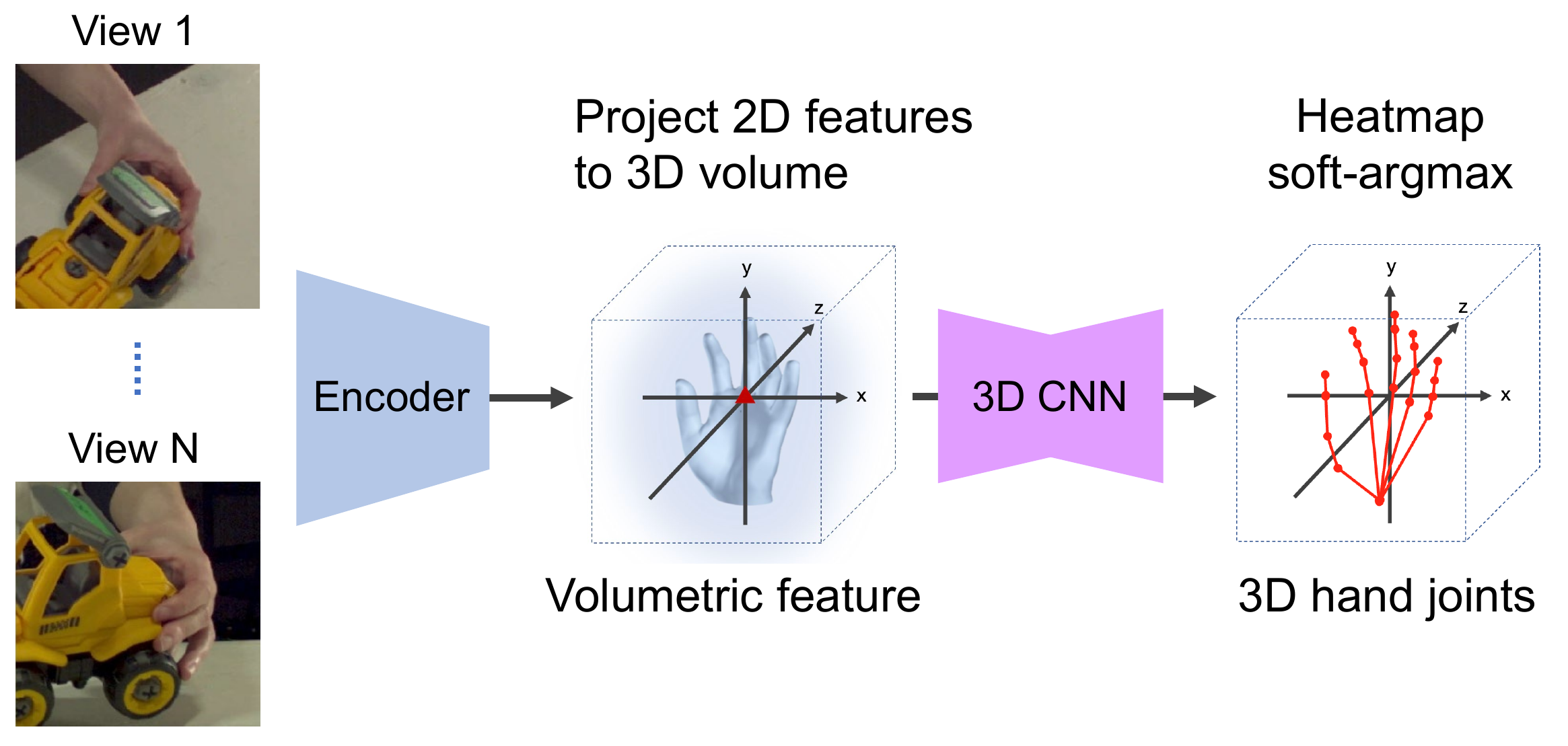}
    \caption{\tbf{Architecture of the hand pose annotation model.} 
    We use an EfficientNet encoder~\cite{tan:icml19} to extract 2D features from multi-view images, then aggregate them into a 3D feature volume, and apply volumetric convolution with V2V-Posenet~\cite{moon:cvpr18}.
    We apply soft-argmax to extract hand joint locations from 3D heatmaps.
    }
    \label{fig:volnet}
\end{figure}

\customparagraph{Volumetric annotation network.}
We next design a neural network model for 3D keypoint annotation. 
With multi-camera setups, a standard approach is to triangulate 2D keypoint detections; 
we call this the ``2D + Triangulation" baseline.
For instance, in InterHand2.6M~\cite{moon:eccv20} this approach can achieve an accuracy of 2.78 mm, owing to the high number of cameras (80 to 140).
However, for Assembly101,  2D + Triangulation only achieves {7.97\,mm} given the limited number of 8 RGB cameras (see Table~\ref{table:annot}).
On the other hand,  end-to-end ``learnable triangulation" methods~\cite{iskakov2019learnable,Bartol:CVPR:2022} are known to outperform standard triangulation for human pose estimation in this regime.
We thus adopt this principle and design a multi-view hand pose estimation network based on 3D volumetric feature aggregation.

We name our volumetric network \volnet, and show its design in Fig.~\ref{fig:volnet}.
First, a feature encoder extracts 2D keypoint features for each view.
We then project the features to a single 3D volume, using the softmax-based weighted average proposed in~\cite{iskakov2019learnable}.
Later, an encoder-decoder network based on 3D convolutions refines the volumetric features and outputs 3D heatmaps.
We obtain 3D joint coordinates with soft-argmax operation on the heatmaps.

For the architecture, we use EfficientNet~\cite{tan:icml19} as an encoder to extract compact 2D features before volumetric aggregation, in order to save GPU memory. 
We use V2V-PoseNet~\cite{moon:cvpr18} as the 3D convolutional network.
During training, we generate 2D hand crops by slightly expanding the region enclosing the manually annotated 2D keypoints.
The 3D volume is {300\,mm} long on each side, centered on the bottom of the middle finger (\ie, the third MCP joint).
We also augment the volume's root position by adding random noise 
to each axis, which prevents the model from always predicting the origin of the volume as the third MCP.
At test time, we crop hand regions based on the output of a hand detector, and use the predicted third MCP from the 2D + Triangulation baseline as the volume root.

\begin{figure*}[th]
    \centering
    \includegraphics[width=.8\linewidth]{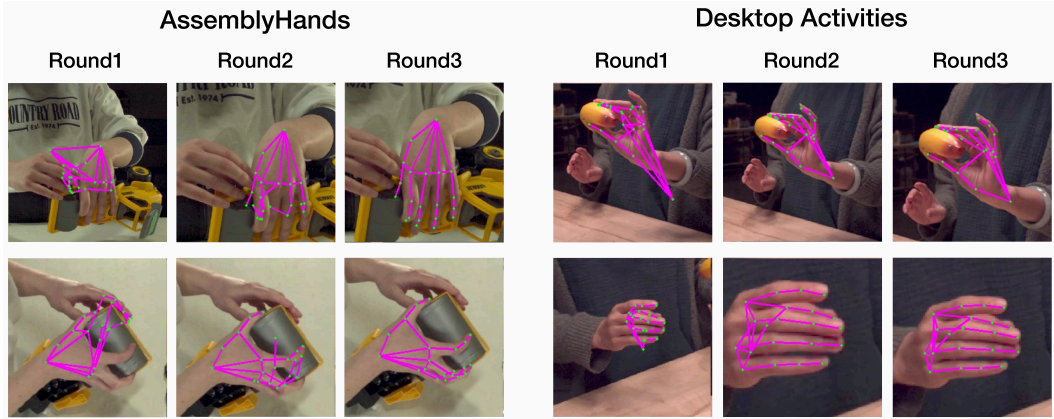}
    \caption{
    \tbf{Example visualization of iterative refinement on AssemblyHands and Desktop Activities~\cite{lv_aria:2022}.}
   Over the refinement iterations, the cropped image progressively becomes better centered on the hand, and the predicted hand pose becomes more accurate.
    }
    \label{fig:refine}
\end{figure*}
\customparagraph{Iterative refinement.}
During the inference of \volnet, we propose a simple iterative refinement heuristic that improves the model's input over several rounds.
As mentioned above, \volnet{} requires hand bounding boxes to crop input images and the root position to construct the 3D volume.
At test time, the bounding box and volume root come from a hand detector and triangulation of initial 2D keypoint predictions, respectively, which may contain inaccuracies.
We found that \volnet{} performs worse than the hypothetical upper bound of having the manually annotated crops and root positions as input.

Our iterative refinement is motivated by this observation: since \volnet{} already generates reasonable predictions, we can use its output to re-initialize the hand crops and volume root position.
This gives the network better inputs with each successive round.
We call the original model \volnet-R1 (the first round of inference), and name the following rounds as \volnet-R2, etc.
In each additional round, we define input hand crops from projected 2D keypoints generated by the \volnet in the previous round, and center the 3D volume on the predicted root position.
Note that we freeze \volnet{} during the iterative refinement inference and only update the input (\ie, bounding box and volume root) to the model.

\subsection{Evaluation of annotated 3D hand poses}
We now compare the accuracy of our proposed annotation method to several baselines, including egocentric hand tracker~\cite{dpe} used in the original Assembly101.
First, to evaluate in-distribution generalization, we use the manually annotated test set from Assembly101, which contains frames sampled from 8 sequences at 1~Hz. 
We also consider the generalization to unseen multi-camera setups; for this purpose, we use the \emph{Desktop Activities} subset from the recently released Aria Pilot Dataset~\cite{lv_aria:2022}; see our supplementary material for the illustration of the camera setup.

\tblannot

\customparagraph{Comparison to egocentric hand pose annotation.}
We compare the accuracy of annotation methods on a manually-annotated evaluation set in Table~\ref{table:annot}.
The original hand annotations in Assembly101~\cite{sener:cvpr22} are computed by an egocentric hand pose estimator, UmeTrack~\cite{dpe}, using monochrome images from egocentric cameras.
The egocentric annotation (Egocentric-only) achieved a error of 27.55mm, which is significant higher than methods using exocentric cameras, namely 2D + Triangulation and our proposed method.
We found that the annotation from egocentric cameras becomes inaccurate when in-hand objects block the user's perspective.
For these cases, the keypoint predictions from multiple exocentric cameras help localize the occluded keypoints.
By fusing volumetric features from multi-view exocentric images, our \volnet{} performs much better than the standard  2D + Triangulation baseline.

\customparagraph{Ablation study of \volnet{}.}
As shown in Table~\ref{table:annot}, our initial inference result (\volnet-R1) achieved reasonable performance with 5.42~mm error.
The iterative refinement further boosts in reducing annotation errors from 5.42~mm to 4.20~mm (22.5\% reduction) after two rounds.

In Fig.~\ref{fig:refine}, we visualize the transition of the hand crops and \volnet's predictions over the rounds on both Assembly101 and Desktop Activities.
At the beginning, hand crops in the first round are not optimal for both datasets.
For example, the model cannot distinguish which hand to annotate because both hands are centered on the image in Assembly101 (left).
Also, the hand moves above in the image (top right) and appears to be tiny (bottom right).
Given these suboptimal hand crops, the prediction becomes noisy, such as keypoint predictions going to the other hand and detaching from the hand position.
However, in the later rounds, the hand crops gradually focus on the target hand (\eg, left hand on the top left figure), which improves the keypoint localization.

\tblegohoi

\customparagraph{Generalization to novel camera configurations.}
To evaluate the cross-dataset generalization ability of our annotation method, 
we use the Desktop Activities dataset, which also features hand-object interactions in a multi-camera setup.
It is recorded with a multi-view camera rig similar to that of Assembly101, but with 12 exocentric RGB cameras and different camera placements.
The objects are from the YCB benchmark~\cite{chao:cvpr21}, which are also unseen in Assembly101. 
To our knowledge, there are no existing hand pose annotations for Desktop Activities.
We use the same manual annotation approach to construct an evaluation set with 1105 annotated frames from three different sequences.

As shown in Table~\ref{tbl:egohoi}, due to the new camera configuration and the presence of novel objects, all methods obtain higher errors than in the Assembly101 setting.
In particular, the baseline annotation method 2D + Triangulation degrades significantly when applied to Desktop Activities, to nearly 50 mm MPJPE.
In contrast, our \volnet{} is quite robust to the new setting, achieving an initial MPJPE of 21.20~mm, and 13.38~mm after two rounds of iterative refinement (a 36.9\% error reduction).

%%%%%%%%%%%%%%%%%%%%%%%%%%%%%%%%%%%%%%%%%%%%%%%%%%%%%%%%%%%%%%%
%%%%%%%%%%%%%%%%%%%%%%%%%%%%%%%%%%%%%%%%%%%%%%%%%%%%%%%%%%%%%%%
\definecolor{green}{RGB}{0,240,0}
\begin{figure*}[t]
    \centering
    \includegraphics[width=.95\linewidth]{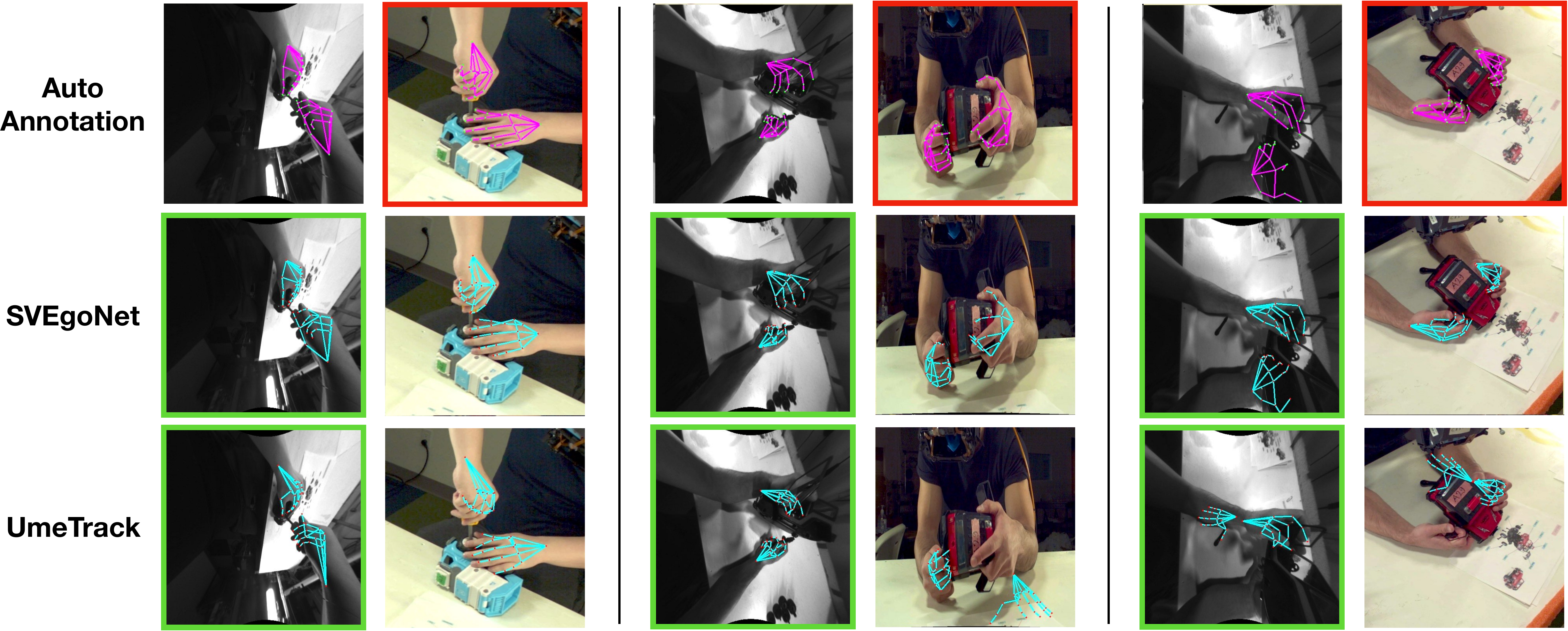}
    \caption{
    \tbf{Qualitative examples of 3D hand poses given by our automatic annotation, \svnet, and UmeTrack~\cite{dpe}.} 
    We visualize the 2D projection of 3D poses in one egocentric image and another synchronized exocentric image.
    We use colored borders to indicate the source images from which hand poses are computed: 
    {\color{red}exocentric (red, additional views omitted)} or {\green{egocentric (green)}}.
    The egocentric-based UmeTrack exhibits multiple failure modes, such as inaccurate relative depth prediction of keypoints (left) and entire hand (middle), and completely losing track during occlusion (right). 
    Our multi-view automatic annotation overcomes these failures, 
    resulting in a more robust \svnet{} when trained on such annotations.
    }
    \label{fig:qual_ego}
\end{figure*}

\section{Egocentric 3D hand pose estimation}\label{sec:egohpe}
To build hand pose estimators for egocentric views, we train models on egocentric images with keypoint annotations generated in Section~\ref{sec:dataset}.
Training on egocentric images is necessary because existing exocentric datasets do not fully capture egocentric-specific biases in terms of the viewpoint, camera characteristics (egocentric cameras are typically fisheye), and blur from the head motion.
Hence, the generalization of exocentric models to egocentric data tends to be limited: for example, in~\cite{ohkawa:eccv22}, the model trained on DexYCB~\cite{chao:cvpr21} (exocentric) achieves 14\% PCK on FPHA~\cite{hernando:cvpr18} (egocentric), compared to 63\% when fine-tuned on FPHA.

\noindent\textbf{Problem setting.}
We conduct an evaluation of a 3D hand pose estimator trained by egocentric images.
Given a single egocentric image, the model aims to predict the 3D coordinates of 21 joints in the wrist-relative space.
We split both the manually annotated and the automatically annotated datasets (M/A) into training and evaluation.
Manually annotated training and evaluation sets contain 19.2K and 3.0K images, respectively, which are sampled at 1\,Hz from 62 video sequences with 14 subjects.
Automatically annotated sets include 405K and 63K images, respectively, which are sampled at \fpsar\,Hz from a disjoint set of 20 sequences with 20 subjects.

\noindent\textbf{Single-view baseline.}
Following standard heatmap-based hand pose estimators~\cite{moon:eccv20,iqbal:eccv18}, 
we build a single-view network (\svnet) trained on monochrome egocentric images.
The model consists of 2.5D heatmap optimization and hand identity classification. 
The 2.5D heatmaps represent 2D keypoint heatmaps in x-y axis and the wrist-relative distance from the camera in z axis.
We use the ResNet-50~\cite{he:cvpr16} backbone.
The 3D joint coordinates are computed by applying the argmax operation on the 2.5D heatmaps.

In addition, we observe that learning the correlations between hand poses and the identity of hand is effective in our task.
For instance, during the ``screw'' motion, right-handed participants in Assembly101 are more likely to hold the toy with their left hand and turn the screwdriver with their right hand.
In another example, when handling small parts, both hands tend to be closer and appear in the same hand crop.
To capture such correlations, we add a hand identity classification branch to \svnet, inspired by~\cite{moon:eccv20}.
We let the branch classify whether \emph{left}, \emph{right}, or \emph{both} hands appear in a given hand crop.

\noindent\textbf{Evaluation.}
We compare the predictions from our model and UmeTrack~\cite{dpe} with the ground truth in wrist-relative coordinates.
We use two standard metrics: mean per joint position error (MPJPE) in millimeters, 
and area under curve of percentage of correct keypoints (PCK-AUC).

\tblcross
\subsection{Results}
\customparagraph{Effect of automatic annotation.}
In Table~\ref{table:cross}, we compare the performance of \svnet{} trained on datasets with manual (M), automatic (A), and manual + automatic (M+A) annotations, respectively.
We provide Eval-M results as the canonical reference and the other results on all evaluation sets.
We observe that using Train-A alone, which is 21 times larger than Train-M, slightly increases error on {Eval-M} by 3\% relative.
On the other hand, the model trained on the combined annotations, Train-M+A, consistently gives the lowest error, which validates our efforts in scaling annotations with automatic methods.
This study also shows that having a hybrid of manual and automatic annotations is a pragmatic solution to improving the model performance.

\customparagraph{Qualitative results.}
Fig.~\ref{fig:qual_ego} shows qualitative examples of 3D hand poses generated by UmeTrack~\cite{dpe}, our automatic annotation pipeline, and our trained egocentric baseline \svnet.
We visualize the prediction of each model from different viewpoints.
The egocentric baseline UmeTrack can estimate hand poses reasonably well when seen from the egocentric view; however, visualization in exocentric views reveals that it tends to make errors along the z-axis.
In particular, the accuracy of the prediction degrades in hard examples with self-occlusion (left example) or hand-object occlusion (middle and right examples).
On the other hand, our multi-view automatic annotation overcomes these failures using the cues of multiple exocentric images.
Owing to it, the \svnet{} trained on the annotation achieves more robust results to these occlusion cases.

%%%%%%%%%%%%%%%%%%%%%%%%%%%%%%%%%%%%%%%%%%%%%%%%%%%%%%%%%%%%%%%
%%%%%%%%%%%%%%%%%%%%%%%%%%%%%%%%%%%%%%%%%%%%%%%%%%%%%%%%%%%%%%%
\section{Action classification from 3D hand poses}\label{sec:action}

\begin{figure*}[t]
   \centering
 {\includegraphics[width=0.30\linewidth]{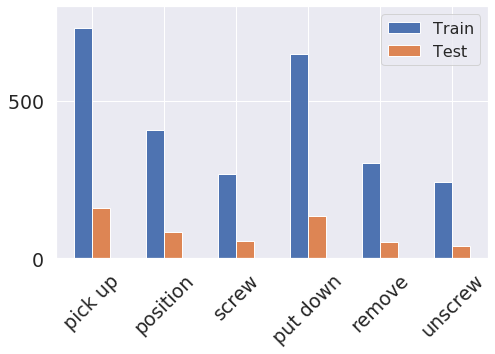}}
    \centering
   \includegraphics[width=0.32\linewidth]{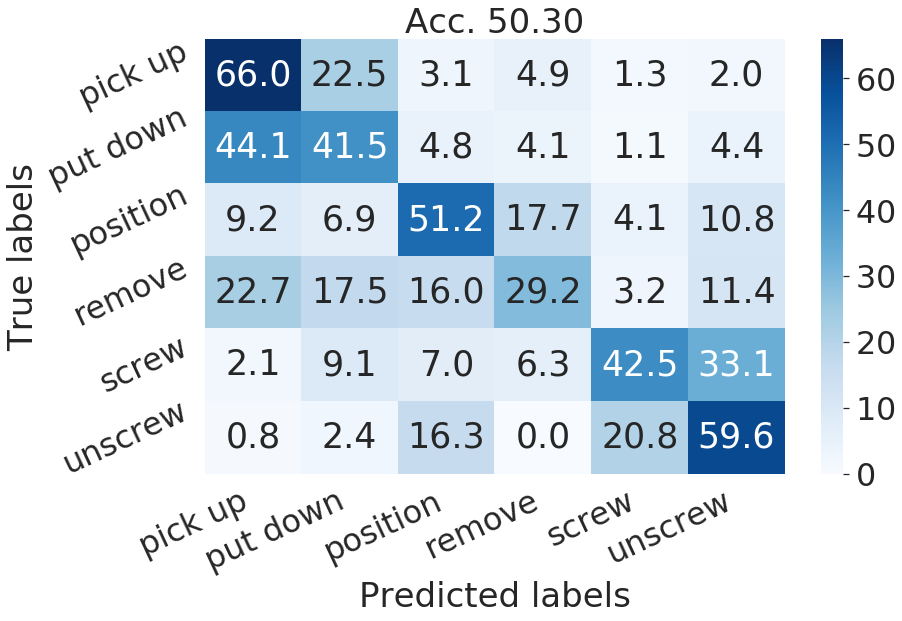}
   \includegraphics[width=0.32\linewidth]{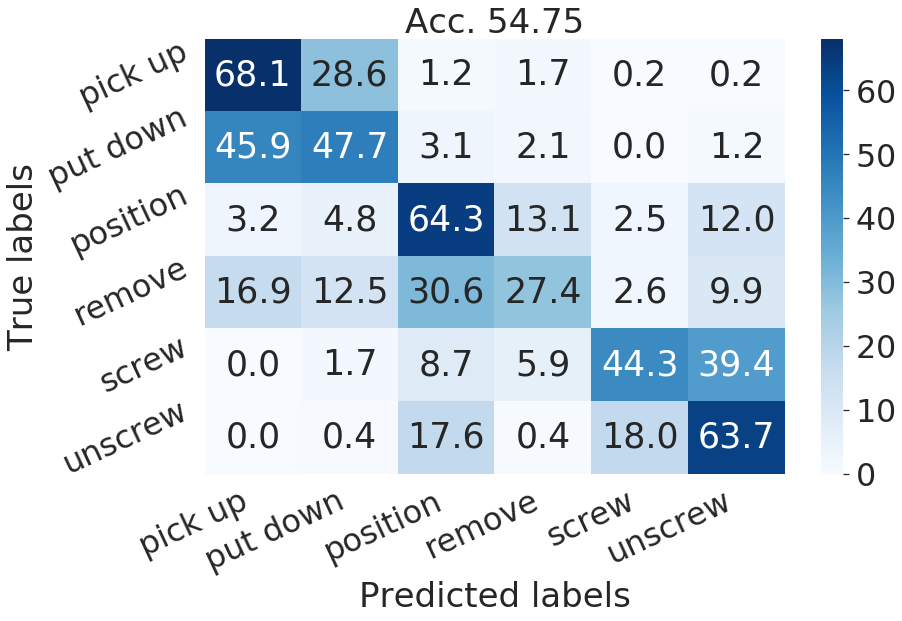}
   \vspace{-.5em}
   \caption{\tbf{Verb label distribution and confusion matrices of verb classification.}
   We show the distribution of the six verb labels (left) used in our experiments and confusion matrices of UmeTrack~\cite{dpe} (middle) and our \svnet{} (right). 
   }
   \label{fig:confmat}
\end{figure*}

Finally, we revisit our motivating question:
\emph{How does the quality of 3D hand poses affect action recognition performance?}
We answer this question with a novel evaluation scheme: verb classification with hand poses as input.
In Assembly101~\cite{sener:cvpr22}, an action is defined at a fine-grained level as the combination of a single verb describing a movement plus an interacting object, \eg, \emph{pick up a screwdriver}. 
We use six verb labels to evaluate predicted hand poses, including \emph{pick up}, \emph{position}, \emph{screw}, \emph{put down}, \emph{remove}, and \emph{unscrew} (see the left figure in Fig.~\ref{fig:confmat}).
This is because these verbs frequently appear in the dataset and heavily depend on the user's hand movements, which hand pose estimation aims to encode.

For classifying verbs, we train MS-G3D~\cite{liu2020disentangling}, a graph convolutional network, using the output of egocentric hand pose estimators.
We note that verbs like \emph{screw} and \emph{unscrew} are cyclic actions that usually take a long time but have relatively fewer instances (see Fig.~\ref{fig:confmat}). 
To address this, we augment the training data for these verb classes.
Following the experiments of Assembly101, for each segment, we input the sequence of 42 keypoints (21 for each hand).
We use the same train/eval split as our automatic annotation, AssemblyHands-A, sampled at a frequency of \fpsar\,Hz (\vs the original 60\,Hz).
The model constructs time-series graphs from 3D hand poses, and classifies each segment of poses into a single verb.

\tblego

\subsection{Results}
In Table~\ref{table:ego}, we report the verb classification accuracy given 3D hand poses estimated from the egocentric cameras. 
First, we establish an empirical upper bound for verb classification accuracy in {AssemblyHands-A} using the automatically annotated hand poses.
The verb classifier trained on our automatic annotations achieves 60\% verb accuracy on average.

We compare our single-view \svnet{} to the off-the-shelf egocentric hand pose estimator UmeTrack~\cite{dpe}, which was used to provide the original annotations for Assembly101, and uses a feature fusion module from multiple egocentric images.
First, we report on the pose estimation metric, where \svnet{} achieves 21.92~mm MPJPE, which is 33\% lower than UmeTrack.
Next, for verb classification accuracy, using hand poses predicted by \svnet{} also outperforms using UmeTrack by a large margin (54.7 \vs 50.3). 
When using the upper bound performance of 60.0 as a reference, 
using \svnet{} poses attains 91.1\% relative performance, 
which is significantly better than the 83.8\% that can be achieved with UmeTrack.

Additionally, we present classification confusion matrices for UmeTrack and \svnet{} in Fig.~\ref{fig:confmat}.
Using \svnet{} predictions significantly reduces the off-diagonal confusions.
Measuring the performance individually per verb, \svnet{} improves the verb accuracy from the UmeTrack by 2.1\%, 6.2\%, 13.1\%, 1.8\%, and 4.1\% for \emph{pick up}, \emph{put down}, \emph{position}, \emph{screw}, and \emph{unscrew}, respectively, while dropping the accuracy for \emph{remove} by 1.8\%.

The fact that we achieve more than 90\% relative performance compared to the upper bound is very encouraging, as \svnet{} only uses a single egocentric image as input, as opposed to performing complex inference with multi-view exocentric images.
This again speaks to the large potential in recognizing activities using lightweight egocentric setups, such as head-mounted monochrome cameras.

%%%%%%%%%%%%%%%%%%%%%%%%%%%%%%%%%%%%%%%%%%%%%%%%%%%%%%%%%%%%%%%
%%%%%%%%%%%%%%%%%%%%%%%%%%%%%%%%%%%%%%%%%%%%%%%%%%%%%%%%%%%%%%%
\section{Conclusion}
\label{sec:conclusion}
We present \tbf{AssemblyHands}, a novel benchmark dataset for studying egocentric activities in the presence of strong hand-object interactions.  
We provide accurate 3D hand pose annotations on a large scale, using an automatic annotation method based on multi-view feature aggregation, which far outperforms the egocentric-based annotation from the original Assembly101.
The accurate annotations allow us to carry out in-depth analysis of how hand pose estimates inform action recognition.
We provide a baseline for single-view egocentric hand pose estimation, and propose a novel evaluation scheme based on verb classification.
Our results have confirmed that the quality of 3D hand poses significantly affects verb recognition performance.
We hope that AssemblyHands can inspire new methods and insights for understanding human activities from the egocentric view.

\customparagraph{Limitations and future work.} We have focused on hand pose annotations and action classification from hand poses in this work.
While object cues (\eg, object pose) would further benefit the task, its annotation creates a bigger challenge due to the presence of many small object parts in the assembly task. 
In future work, we first plan to extend hand pose annotation to the entire Assembly101 at higher sampling rates.
We also plan to obtain object-level annotation, \eg, object bounding boxes.
Finally, we are interested in exploring the interplay between hands, objects, and actions with multi-task learning.

\subsection*{Acknowledgments}
The authors would like to thank Kevin Harris for help with data collection, and Lingni Ma, Svetoslav Kolev, Bugra Tekin, Edoardo Remelli, Shangchen Han, Robert Wang for helpful discussions.

%%%%%%%%%%%%%%%%%%%%%%%%%%%%%%%%%%%%%%%%%%%%%%%%%%%%%%%%%%%%%%%
%%%%%%%%%%%%%%%%%%%%%%%%%%%%%%%%%%%%%%%%%%%%%%%%%%%%%%%%%%%%%%%
%%%%%%%%% REFERENCES
%%%%%%%%%%%%%%%%%%%%%%%%%%%%%%%%%%%%%%%%%%%%%%%%%%%%%%%%%%%%%%%
%%%%%%%%%%%%%%%%%%%%%%%%%%%%%%%%%%%%%%%%%%%%%%%%%%%%%%%%%%%%%%%
{\small
\bibliographystyle{ieee_fullname}
\bibliography{ref_base,ref_hands,ref_behavior,egbib}

\begin{thebibliography}{10}\itemsep=-1pt

\bibitem{Bartol:CVPR:2022}
K. Bartol, D. Bojani\'{c}, T. Petkovi\'{c}, and T. Pribani\'{c}.
\newblock Generalizable human pose triangulation.
\newblock In {\em \cvpr}, 2022.

\bibitem{cao:tpami2019}
Z. {Cao}, G. {Hidalgo Martinez}, T. {Simon}, S. {Wei}, and Y.~A. {Sheikh}.
\newblock Openpose: Realtime multi-person 2d pose estimation using part
  affinity fields.
\newblock {\em IEEE Transactions on Pattern Analysis and Machine Intelligence},
  2019.

\bibitem{chao:cvpr21}
Y.-W. Chao, W. Yang, Y. Xiang, P. Molchanov, A. Handa, J. Tremblay, Y.~S.
  Narang, K. {Van~Wyk}, U. Iqbal, S. Birchfield, J. Kautz, and D. Fox.
\newblock {DexYCB}: A benchmark for capturing hand grasping of objects.
\newblock In {\em \cvpr}, pages 9044--9053, 2021.

\bibitem{cheron:2015iccv}
G. Ch{\'e}ron, I. Laptev, and C. Schmid.
\newblock P-cnn: Pose-based cnn features for action recognition.
\newblock In {\em \iccv}, pages 3218--3226, 2015.

\bibitem{damen:ijcv21}
D. Damen, H. Doughty, G.~M. Farinella, A. Furnari, J. Ma, E. Kazakos, D.
  Moltisanti, J. Munro, T. Perrett, W. Price, and M. Wray.
\newblock Rescaling egocentric vision.
\newblock {\em International Journal of Computer Vision (IJCV)}, early access,
  2021.

\bibitem{feng:arxiv21}
Q. Feng, K. He, H. Wen, C. Keskin, and Y. Ye.
\newblock Active learning with pseudo-labels for multi-view 3d pose estimation.
\newblock {\em CoRR}, abs/2112.13709, 2021.

\bibitem{hernando:cvpr18}
G. Garcia-Hernando, S. Yuan, S. Baek, and T.-K. Kim.
\newblock First-person hand action benchmark with {RGB-D} videos and {3D} hand
  pose annotations.
\newblock In {\em \cvpr}, pages 409--419, 2018.

\bibitem{grauman:cvpr22}
K. Grauman, A. Westbury, E. Byrne, Z. Chavis, A. Furnari, R. Girdhar, J.
  Hamburger, H. Jiang, M. Liu, X. Liu, M. Martin, T. Nagarajan, I. Radosavovic,
  S.~K. Ramakrishnan, F. Ryan, J. Sharma, M. Wray, M.g Xu, E.~Zhongcong Xu, C.
  Zhao, S. Bansal, D. Batra, V. Cartillier, S. Crane, T. Do, M. Doulaty, A.
  Erapalli, C. Feichtenhofer, A. Fragomeni, Q. Fu, C. Fuegen, A. Gebreselasie,
  C. Gonzalez, J. Hillis, X. Huang, Y. Huang, W. Jia, W. Khoo, J. Kolar, S.
  Kottur, A. Kumar, F. Landini, C. Li, Y. Li, Z. Li, K. Mangalam, R. Modhugu,
  J. Munro, T. Murrell, T. Nishiyasu, W. Price, P.~R. Puentes, M. Ramazanova,
  L. Sari, K. Somasundaram, A. Southerland, Y. Sugano, R. Tao, M. Vo, Y. Wang,
  X. Wu, T. Yagi, Y. Zhu, P. Arbelaez, D. Crandall, D. Damen, G.~M. Farinella,
  B. Ghanem, V.~K. Ithapu, C.~V. Jawahar, H. Joo, K. Kitani, H. Li, R.
  Newcombe, A. Oliva, H.~Soo Park, J.~M. Rehg, Y. Sato, J. Shi, M.~Z. Shou, A.
  Torralba, Lo Torresani, M.i Yan, and J. Malik.
\newblock {Ego4D}: Around the world in 3, 000 hours of egocentric video.
\newblock In {\em \cvpr}, pages 18973--18990, 2022.

\bibitem{hampali:cvpr20}
S. Hampali, M. Rad, M. Oberweger, and V. Lepetit.
\newblock Honnotate: A method for {3D} annotation of hand and object poses.
\newblock In {\em \cvpr}, pages 3196--3206, 2020.

\bibitem{megatrack}
S. Han, B. Liu, R. Cabezas, C. Twigg, P. Zhang, J. Petkau, T.-H. Yu, C.-J. Tai,
  M. Akbay, Z. Wang, A. Nitzan, G. Dong, Y. Ye, L. Tao, C. Wan, and R. Wang.
\newblock {MEgATrack}: monochrome egocentric articulated hand-tracking for
  virtual reality.
\newblock {\em ACM Transactions on Graphics (ToG)}, 39(4):87--1, 2020.

\bibitem{dpe}
S. Han, P.-C. Wu, Y. Zhang, B. Liu, L. Zhang, Z. Wang, W. Si, P. Zhang, Y. Cai,
  T. Hodan, R. Cabezas, L. Tran, M. Akbay, T.-H. Yu, C. Keskin, and R. Wang.
\newblock {UmeTrack}: Unified multi-view end-to-end hand tracking for {VR}.
\newblock In {\em \sga}, pages 50:1--50:9, 2022.

\bibitem{hasson:cvpr19}
Y. Hasson, G. Varol, D. Tzionas, I. Kalevatykh, M.~J. Black, I. Laptev, and C.
  Schmid.
\newblock Learning joint reconstruction of hands and manipulated objects.
\newblock In {\em \cvpr}, pages 11807--11816, 2019.

\bibitem{he:cvpr16}
K. He, X. Zhang, S. Ren, and J. Sun.
\newblock Deep residual learning for image recognition.
\newblock In {\em \cvpr}, pages 770--778, 2016.

\bibitem{iqbal:2017}
U. Iqbal, M. Garbade, and J. Gall.
\newblock Pose for action - action for pose.
\newblock In {\em Proceedings of the {IEEE} International Conference on
  Automatic Face {\&} Gesture Recognition, {FG}}, pages 438--445, 2017.

\bibitem{iqbal:eccv18}
U. Iqbal, P. Molchanov, T.~M. Breuel, J. Gall, and J. Kautz.
\newblock Hand pose estimation via latent {2.5D} heatmap regression.
\newblock In {\em \eccv}, pages 125--143, 2018.

\bibitem{iskakov2019learnable}
K. Iskakov, E. Burkov, V. Lempitsky, and Y. Malkov.
\newblock Learnable triangulation of human pose.
\newblock In {\em \iccv}, pages 7718--7727, 2019.

\bibitem{kong:2022ijcv}
Y. Kong and Y. Fu.
\newblock Human action recognition and prediction: A survey.
\newblock {\em International Journal of Computer Vision}, 130(5):1366--1401,
  2022.

\bibitem{kwon:iccv21}
T. Kwon, B. Tekin, J. St{\"{u}}hmer, F. Bogo, and M. Pollefeys.
\newblock {H2O:} two hands manipulating objects for first person interaction
  recognition.
\newblock In {\em \iccv}, pages 10118--10128, 2021.

\bibitem{lin2019tsm}
J. Lin, C. Gan, and S. Han.
\newblock {TSM}: Temporal shift module for efficient video understanding.
\newblock In {\em \iccv}, pages 7083--7093, 2019.

\bibitem{liu2020disentangling}
Z. Liu, H. Zhang, Z. Chen, Z. Wang, and W. Ouyang.
\newblock Disentangling and unifying graph convolutions for skeleton-based
  action recognition.
\newblock In {\em \cvpr}, pages 143--152, 2020.

\bibitem{lv_aria:2022}
Z. Lv, E. Miller, J. Meissner, L. Pesqueira, C. Sweeney, J. Dong, L. Ma, P.
  Patel, P. Moulon, K. Somasundaram, O. Parkhi, Y. Zou, N. Raina, S. Saarinen,
  Y.~M. Mansour, P.-K. Huang, Z. Wang, A. Troynikov, R.~M. Artal, D. DeTone, D.
  Barnes, E. Argall, A. Lobanovskiy, D.~J. Kim, P. Bouttefroy, J. Straub, J.~J.
  Engel, P. Gupta, M. Yan, R.~D. Nardi, and R. Newcombe.
\newblock Aria pilot dataset.
\newblock \url{https://about.facebook.com/realitylabs/projectaria/datasets},
  2022.

\bibitem{moon:cvpr18}
G. Moon, J.~Y. Chang, and K.~M. Lee.
\newblock {V2V-PoseNet}: Voxel-to-voxel prediction network for accurate {3D}
  hand and human pose estimation from a single depth map.
\newblock In {\em \cvpr}, pages 5079--5088, 2018.

\bibitem{moon:eccv20}
G. Moon, S.-I. Yu, H. Wen, T. Shiratori, and K.~M. Lee.
\newblock {InterHand2.6M}: A dataset and baseline for {3D} interacting hand
  pose estimation from a single {RGB} image.
\newblock In {\em \eccv}, pages 548--564, 2020.

\bibitem{mueller:iccv17}
F. Mueller, D. Mehta, O. Sotnychenko, S. Sridhar, D. Casas, and C. Theobalt.
\newblock Real-time hand tracking under occlusion from an egocentric {RGB-D}
  sensor.
\newblock In {\em \iccv}, pages 1163--1172, 2017.

\bibitem{ohkawa:arxiv22}
T. Ohkawa, R. Furuta, and Y. Sato.
\newblock Efficient annotation and learning for 3d hand pose estimation: {A}
  survey.
\newblock {\em CoRR}, abs/2206.02257, 2022.

\bibitem{ohkawa:eccv22}
T. Ohkawa, Y.{-}J. Li, Q. Fu, R. Furuta, K.~M. Kitani, and Y. Sato.
\newblock Domain adaptive hand keypoint and pixel localization in the wild.
\newblock In {\em \eccv}, pages 68–--87, 2022.

\bibitem{romero:tog17}
J. Romero, D. Tzionas, and M.~J. Black.
\newblock Embodied hands: Modeling and capturing hands and bodies together.
\newblock {\em \tog}, 36(6):245:1--245:17, 2017.

\bibitem{sener:cvpr22}
F. Sener, D. Chatterjee, D. Shelepov, K. He, D. Singhania, R. Wang, and A. Yao.
\newblock Assembly101: {A} large-scale multi-view video dataset for
  understanding procedural activities.
\newblock In {\em \cvpr}, pages 21096--21106, 2022.

\bibitem{shahroudy2016ntu}
A. Shahroudy, J. Liu, T.-T. Ng, and G. Wang.
\newblock Ntu rgb+ d: A large scale dataset for 3d human activity analysis.
\newblock In {\em \cvpr}, pages 1010--1019, 2016.

\bibitem{simon:cvpr17}
T. Simon, H. Joo, I. Matthews, and Y. Sheikh.
\newblock Hand keypoint detection in single images using multiview
  bootstrapping.
\newblock In {\em \cvpr}, pages 4645--4653, 2017.

\bibitem{sridhar:eccv16}
S. Sridhar, F. Mueller, M. Zollhoefer, D. Casas, A. Oulasvirta, and C.
  Theobalt.
\newblock Real-time joint tracking of a hand manipulating an object from
  {RGB-D} input.
\newblock In {\em \eccv}, pages 294--310, 2016.

\bibitem{tan:icml19}
M. Tan and Q.~V. Le.
\newblock {EfficientNet}: Rethinking model scaling for convolutional neural
  networks.
\newblock In {\em \icml}, volume~97, pages 6105--6114, 2019.

\bibitem{wang:2013cvpr}
C. Wang, Y. Wang, and A. Yuille.
\newblock An approach to pose-based action recognition.
\newblock In {\em \cvpr}, pages 915--922, 2013.

\bibitem{yao:2011bmvc}
A. Yao, J. Gall, G. Fanelli, and L. Van~Gool.
\newblock Does human action recognition benefit from pose estimation?''.
\newblock In {\em \bmvc}. BMV press, 2011.

\bibitem{yu2017spatio}
B. Yu, H. Yin, and Z. Zhu.
\newblock Spatio-temporal graph convolutional networks: A deep learning
  framework for traffic forecasting.
\newblock {\em arXiv preprint arXiv:1709.04875}, 2017.

\bibitem{zimmermann:iccv17}
C. Zimmermann and T. Brox.
\newblock Learning to estimate {3D} hand pose from single {RGB} images.
\newblock In {\em \iccv}, pages 4913--4921, 2017.

\bibitem{zimmermann:iccv19}
C. Zimmermann, D. Ceylan, J. Yang, B. Russell, M.~J. Argus, and T. Brox.
\newblock {FreiHAND}: A dataset for markerless capture of hand pose and shape
  from single {RGB} images.
\newblock In {\em \iccv}, pages 813--822, 2019.

\end{thebibliography}
}

%%%%%%%%%%%%%%%%%%%%%%%%%%%%%%%%%%%%%%%%%%%%%%%%%%%%%%%%%%%%%%%
%%%%%%%%%%%%%%%%%%%%%%%%%%%%%%%%%%%%%%%%%%%%%%%%%%%%%%%%%%%%%%%
%%%%%%%%% Appendix
%%%%%%%%%%%%%%%%%%%%%%%%%%%%%%%%%%%%%%%%%%%%%%%%%%%%%%%%%%%%%%%
%%%%%%%%%%%%%%%%%%%%%%%%%%%%%%%%%%%%%%%%%%%%%%%%%%%%%%%%%%%%%%%
%\pagebreak
\appendix
\section*{Appendix}

\section{Camera rigs}
% https://tex.stackexchange.com/questions/55764/input-a-figure-between-title-and-body-in-twocolumn-form
%\twocolumn[{%
%\renewcommand\twocolumn[1][]{#1}%
%\maketitle
\begin{figure*}[!ht]
\begin{center}
\vspace{-1em}
    \centering
    \captionsetup{type=figure}
    \includegraphics[width=0.448\linewidth]{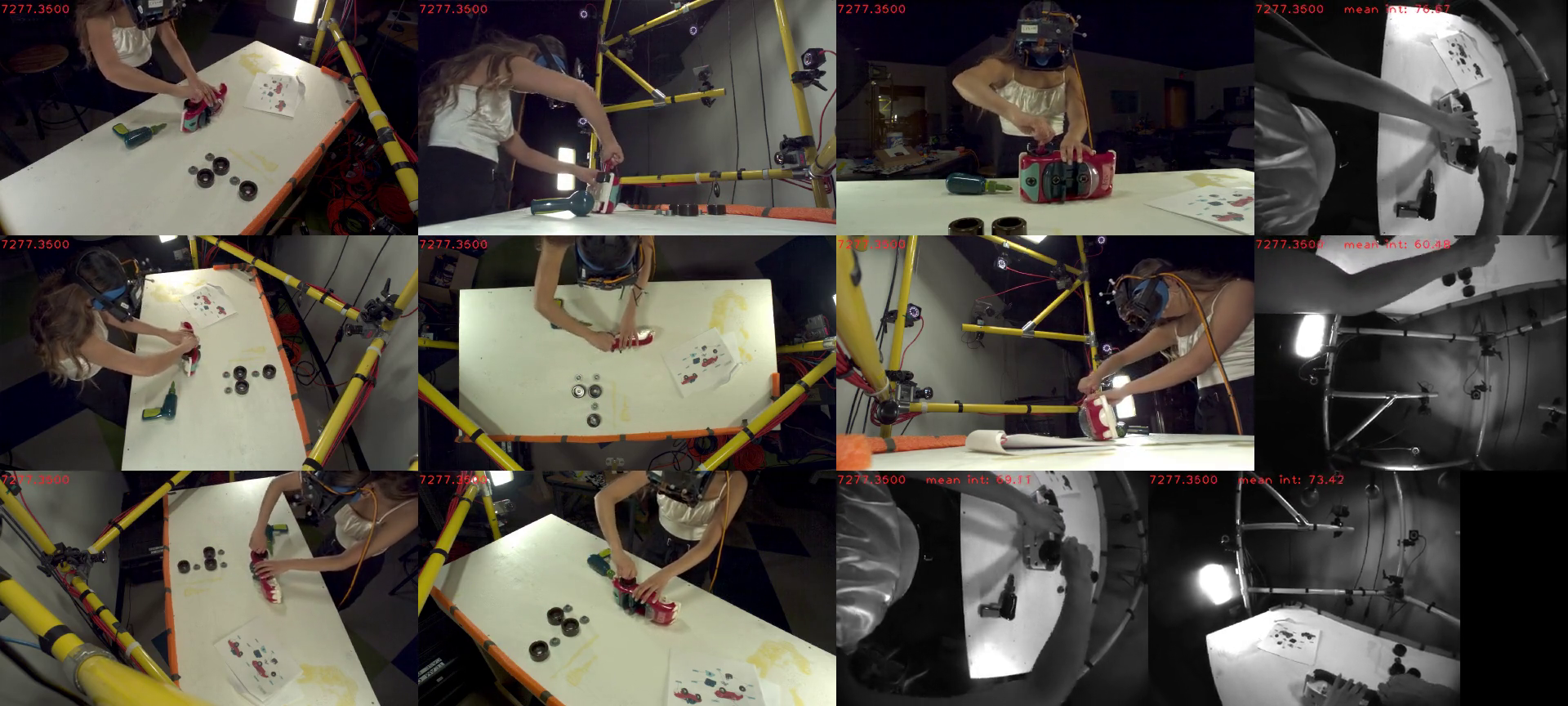}
    \includegraphics[width=0.546\linewidth]{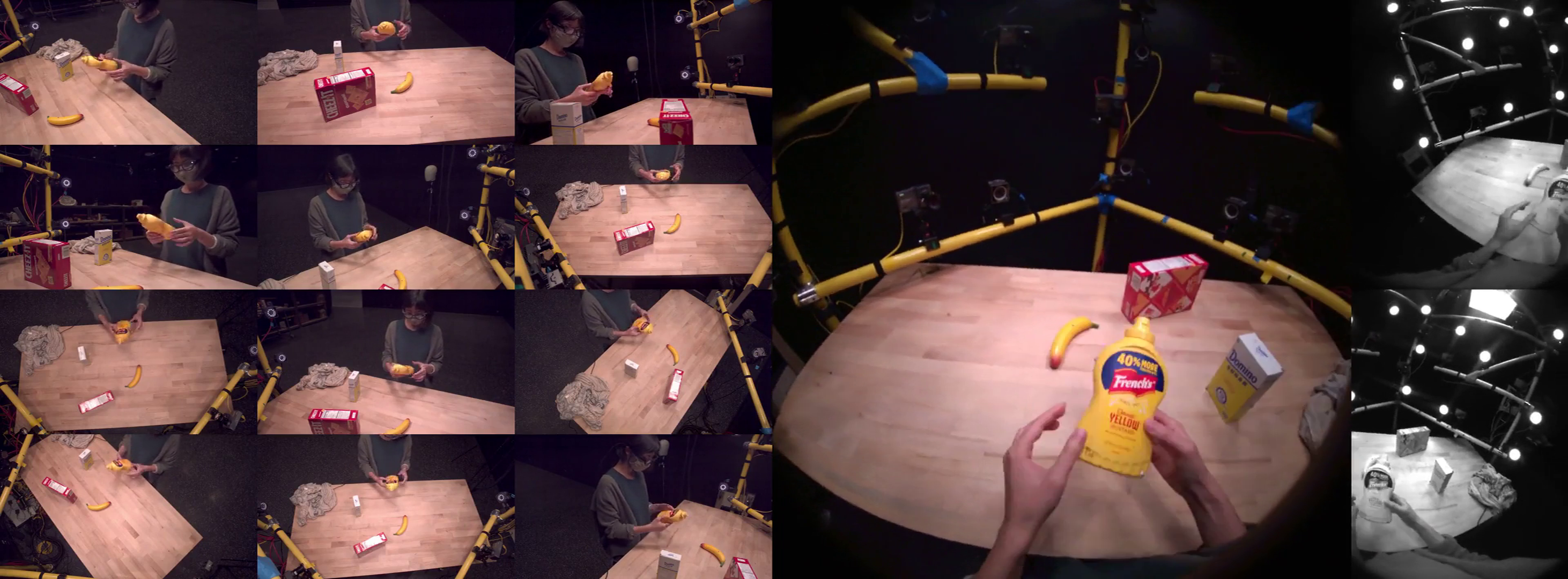}
    \captionof{figure}{
    \textbf{Multi-view camera rigs used in our experiments.} Left: Assembly101. Right: Desktop Activities.
    }
    \label{fig:cam_config}
%\vspace{1em}
\end{center}%
\end{figure*}
%}]

%%%%%%%%%%%%%%%%%%%%%%%%%%%%%%%%%%%%%%%%%%%%%%%%%%%%%%%%%%%%%%%%%%%%%%%%%%%
%%%%%%%%%%%%%%%%%%%%%%%%%%%%%%%%%%%%%%%%%%%%%%%%%%%%%%%%%%%%%%%%%%%%%%%%%%%
Please see example captured frames from both camera rigs used in our experiments in Fig.~\ref{fig:cam_config}.

AssemblyHands uses the same camera rig as Assembly101~\cite{sener:cvpr22}, which consists of 8 RGB cameras of 1080p resolution mounted on a static scaffold, and a synchronized headset with 4 monochrome cameras of VGA resolution, arranged similarly to the Oculus Quest VR headset.

We also use  another multi-camera setup from the \emph{Desktop Activities} subset in the recent Aria Pilot Dataset~\cite{lv_aria:2022}, which has 12 RGB cameras of 1080p resolution, synchronized to the Project Aria glasses.
The glasses are equipped with one egocentric RGB camera and two monochrome cameras, but we only use the static exocentric RGB cameras for the purpose of evaluating our multi-view automatic annotation method.

%%%%%%%%%%%%%%%%%%%%%%%%%%%%%%%%%%%%%%%%%%%%%%%%%%%%%%%%%%%%%%%%%%%%%%%%%%%
%%%%%%%%%%%%%%%%%%%%%%%%%%%%%%%%%%%%%%%%%%%%%%%%%%%%%%%%%%%%%%%%%%%%%%%%%%%
%%%%%%%%%%%%%%%%%%%%%%%%%%%%%%%%%%%%%%%%%%%%%%%%%%%%%%%%%%%%%%%%%%%%%%%%%%%
\section{Implementation details}

\subsection{Processing of exocentric images}
For all training and evaluation images, we pre-process them to remove the lens distortion from both exocentric images. 
This significantly simplifies geometric operations such as triangulation.

At training time, we first project the annotated 3D hand keypoints to each camera's image plane, then use the resulting 2D keypoints to define hand bounding boxes. 
Given a set of keypoint coordinates $\{(x_i,y_i), \forall i\in I\}$ on a single hand (or both hands), we create a 
square bounding box centered on the geometric center of all keypoints, with the side length $L$ defined as
\begin{align}
L= 
\gamma\cdot \max\left(\max_{i\in I}x_i-\min_{i\in I}x_i, \max_{i\in I}y_i-\min_{i\in I}y_i\right),
\end{align}
where $\gamma$ is an expansion coefficient. It is randomized during training with a mean of $1.5$.
For the 3D feature volume, the size is {300\,mm} on each side, centered on the the third MCP joint.
The root position is also augmented during training by adding a small random noise in the range of $[-5~\textrm{mm}, 5~\textrm{mm}]$ to all axes; 
we have observed that without this augmentation,
the network can learn a trivial solution $(0,0,0)$ for the root joint.

At test time, we need to crop hands based on the output of a hand detector. 
However, we found it challenging to directly apply  2D object detectors for hands: off-the-shelf models do not achieve very good accuracy in our setup due to the challenging occlusions, and their detections from different views are not necessarily geometrically consistent.
Instead, we use a more robust heuristic based on the triangulation of body keypoints.
Specifically, we first detect 3D body keypoints from multi-view exocentric images, using the  ``2D + Triangulation'' approach with  a full-image body keypoint detector trained on MS COCO.
Then, we create a virtual ``hand center'' keypoint in 3D, by extending 1/3 of the forearm length (defined from the elbow and wrist keypoints) out from the wrist. 
Finally, for every camera view, we can derive an image-space bounding box, centered on the 2D projection of the ``hand center'', with a heuristic size.
We note that a similar heuristic is employed in the open-source implementation of OpenPose~\cite{cao:tpami2019}.

\begin{figure*}[!ht]
    \centering
    \includegraphics[width=0.8\linewidth]{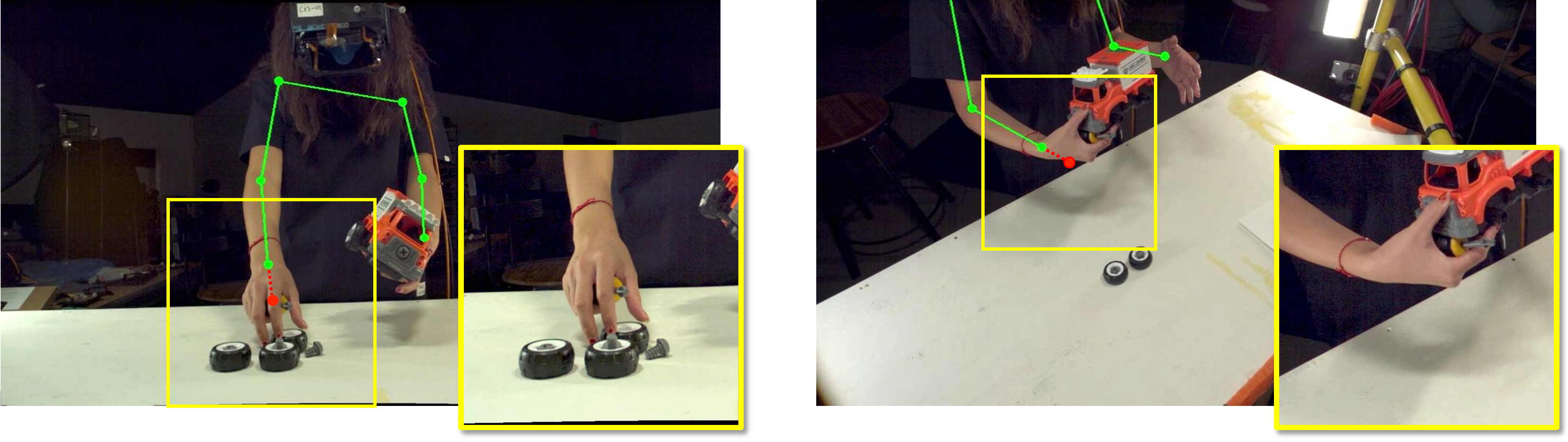}
    \caption{\tbf{Visualization of hand detection on two different frames.}
    Green: detected body keypoints (shown: shoulder, elbow, wrist).
    Red: estimated hand center.
    Yellow: hand bounding boxes derived from our heuristic, and the resulting cropped images.
    Left: the hand is well centered in the bounding box when the wrist is not bent. 
    Right: hand is not centered when the wrist is bent; this can  be corrected by our iterative refinement at inference time.
    }
    \label{fig:hand_detection} 
\end{figure*}

This initialization of hand bounding boxes is crucial for the quality of automatically generated annotations.
We found that our iterative refinement during the annotation pipeline is a key step to reduce the annotation error.
However, when wrong hand boxes are given in the initialization due to misidentifying left or right hand, the refinement further accumulates the error because a new hand crop is generated from the keypoint prediction on the misidentified hand crop.
The introduced hand detection based on 3D body keypoints is more robust to such failures as the hand identity is determined from 
the skeleton structure of human body, not hand image itself.

Fig.~\ref{fig:hand_detection} shows examples of detected body keypoints and derived hand bounding boxes.
By construction, the bounding boxes in different views of the same frame are geometrically consistent, and can be directly used in creating the 3D feature volume.
On the other hand, a drawback of this approach is that  the heuristic guess of ``hand center'' can be inaccurate, especially when the wrist is significantly bent.
As a result, the crop is not necessarily centered on the hand, and we need to use a larger size to ensure coverage.
However, this can be corrected by our iterative refinement scheme during inference; see Fig.~4 in the paper.

\subsection{Processing of egocentric images}
Based on the annotated keypoints, we generate hand clips for input to the egocentric hand pose estimator.
Given the 3D world-space coordinates of all joint locations (21 per hand) on two hands and an egocentric camera, we project the keypoints to the 2D image space, and then crop the image using bounding boxes that enclose the 2D keypoints.
We remove lens distortion from the original fisheye cameras so that the images correspond to simple pinhole cameras.
We also remove crops that are too close to the image boundary or do not contain hands in a given image.
This preprocessing is separately done for each right or left hand, so we have two input crops when the two hands are shown in an image.

Then, given a single cropped image where either the left or right hand appears, we predict the 3D coordinates of 21 joints in the wrist-relative space.
In post-processing, using the predictions for each hand, we convert a single-hand prediction to two-hand prediction by merging two predictions on different hand crops generated from the same single image.
This two-hand format is used for the pose evaluation with the action recognition model.

%%%%%%%%%%%%%%%%%%%%%%%%%%%%%%%%%%%%%%%%%%%%%%%%%%%%%%%%%%%%%%%%%%%%%%%%%%%
%%%%%%%%%%%%%%%%%%%%%%%%%%%%%%%%%%%%%%%%%%%%%%%%%%%%%%%%%%%%%%%%%%%%%%%%%%%
%%%%%%%%%%%%%%%%%%%%%%%%%%%%%%%%%%%%%%%%%%%%%%%%%%%%%%%%%%%%%%%%%%%%%%%%%%%
\section{Annotation quality}

%%%%%%%%%%%%%%%%%%%%%%%%%%%%%%%%%%%%%%%%%%%%%%%%%%%%%%%%%%%%%%%%%%%%%%%%%%%
\subsection{
Comparison to annotation using OpenPose
}

The open-source OpenPose~\cite{cao:tpami2019} has been used to automatically annotate hand poses in several existing datasets, \eg, H2O.
We compare to this annotation approach, by running a 2D + Triangulation baseline with the hand keypoints predicted by OpenPose.

Table~\ref{table:openpose} shows the comparison between the OpenPose baseline and our proposed methods.
First, we note that 2D + Triangulation with OpenPose fails to triangulate on 40.2\% of all the annotated keypoints in our test set,  
due to either too few 2D detections or large triangulation error.
While it does achieve a reasonable 5.15 mm MPJPE on the successfully triangulated predictions, 
the high number of missing annotations is undesirable for the purpose of training an egocentric model.
Note that we vary the distance threshold between 0 and 20~mm when computing PCK-AUC, and we observe OpenPose's PCK value at the maximum cutoff is 59.2\%. 
The rest either do not receive a valid prediction, 
or have a prediction error larger than 20~mm.

Both our 2D + Triangulation baseline and final model \volnet-R3 significantly outperform OpenPose.
First, our 2D + Triangulation model significantly increases the ratio of successful triangulations, at the cost of slightly higher MPJPE.
Then, \volnet-R3 has both the lowest MPJPE and a much higher successful annotation rate, with a PCK value over 85\% at the 20 mm threshold.

\begin{table}[t]
\centering
\begin{tabular}{l|cc}
Annotation method                           & MPJPE & PCK-AUC \\ \hline \hline
2D + Triangulation (OpenPose)               & ~5.15$^*$  & 48.1    \\ 
2D + Triangulation (Ours)                   & 7.97  & 63.8    \\ %\hline
\volnet-R3 (Ours)                           & 4.20  & 83.4      % \\
\end{tabular}
\small{* {Evaluated on valid predictions only.}}
\caption{\textbf{Comparison to OpenPose-based annotation.}
While 2D + Triangulation with OpenPose is relatively accurate on the valid predictions, 
it fails to triangulate on 40.2\% of the keypoints, leading to a very low PCK-AUC value.
We use a maximum distance threshold of 20~mm for PCK-AUC.
Our proposed 2D + Triangulation and  \volnet-R3
both outperform OpenPose by a significant margin.
}
\label{table:openpose}
\end{table}

%%%%%%%%%%%%%%%%%%%%%%%%%%%%%%%%%%%%%%%%%%%%%%%%%%%%%%%%%%%%%%%%%%%%%%%%%%%

\subsection{Verb-wise annotation error}
\begin{figure*}[t]
    \centering
    \includegraphics[width=.6\linewidth]{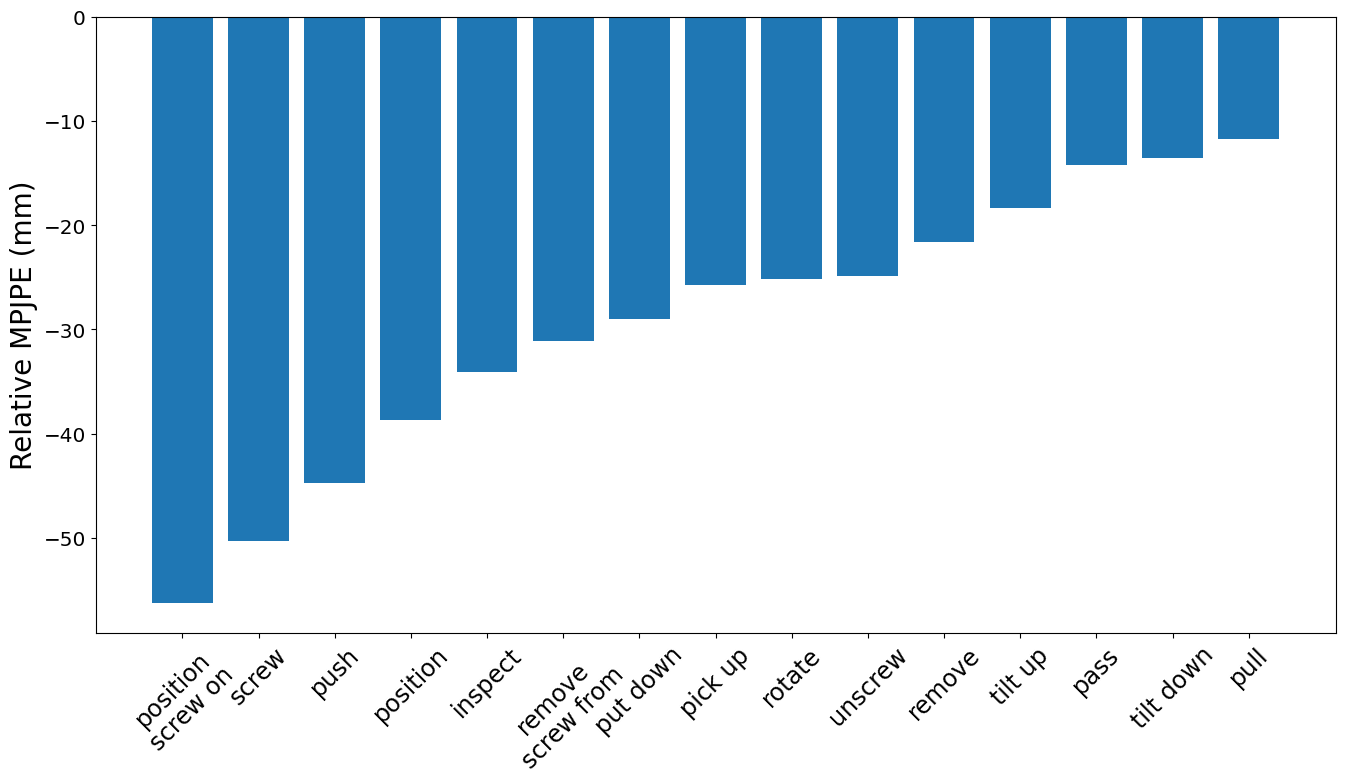}
    \caption{
    \textbf{Verb-wise hand pose annotation error reduction in AssemblyHands.}
    We show relative MPJPE (mm) of our annotations to the original annotations provided by Assembly101~\cite{sener:cvpr22}.
    We observe large reductions in annotation error across all verbs.
    }
    \label{fig:verb_annot}
\end{figure*}

We evaluate our annotation method for each verb category on manually annotated data.
As shown in Fig.~\ref{fig:verb_annot}, we plot relative improvement of our final annotation method (\ie, \volnet{}-R3) to the original annotation based on egocentric hand tracker~\cite{dpe} in Assembly101.
For all the verbs appearing in the manually annotated set, our method further reduces the error by more than 10~mm compared to the original annotation.
Our annotation particularly improves the quality when two hands are interacting with objects intricately, such as \emph{position screw on}, \emph{screw},  \emph{push}, and \emph{position}.
This indicates that our multi-view annotation is more effective during such heavy hand-object occlusion than the annotation from egocentric views only.

Owing to this error reduction in pose annotations, in the downstream task of verb classification (see Table~5 in the paper), the performance of our \svnet{} for \emph{position} is improved by a large margin of 13\%.
Considering the quality of the annotation even for each verb and its support for verb classification, more refined pose annotation helps accurately recognize user's hand actions.

\section{Action recognition}

\subsection{Verb label selection}

In Assembly101~\cite{sener:cvpr22}, fine-grained actions are defined as the combination of a verb and an interacting object, which consists of 24 classes.
Since the verb labels follow a long-tailed distribution, we select the six most frequent verbs out of the full list of 24 for our study.
These include three main assembly verbs: \emph{pick up}, \emph{position}, \emph{screw}, and three disassembly verbs: \emph{put down}, \emph{remove}  and \emph{unscrew}, which altogether cover 70\% of the verb labels introduced in Assembly101.

\subsection{Object cues in verb recognition}
As noted in the future work (Section~6), we believe using object information could further help in recognizing the user's actions.
Since object annotation (\eg, box and pose) for small parts is challenging for this dataset nowadays, we try to use object labels for verb recognition.
We incorporate object class labels as one-hot encoded frame-level features into our pose-based verb classifier.
As we reported in Table~5, the pose-only recognition using \svnet{} has a verb recognition accuracy of 54.7\%.
% when we use automatically generated annotations.
In contrast, the classifier based on pose + object labels achieves a higher accuracy of 56.1\%. %58.5\%.
This improvement further inspires us to explore the use of object bounding boxes and object poses.

\end{document}